# Opinion: Revisiting synthetic data classifications from a privacy perspective

*2025.*

*Authors: Vibeke Binz Vallevik [1, 2], Serena Elizabeth Marshall [2], Aleksandar Babic [2], Jan F. Nygård [3,4]*

## Abstract

Synthetic data is emerging as a cost-effective solution necessary to meet the increasing data demands of AI development, created either from existing knowledge or derived from real data. The traditional classification of synthetic data types into hybrid, partial or fully synthetic datasets has limited value and does not reflect the ever-increasing methods to generate synthetic data. The generation method and their source jointly shape the characteristics of synthetic data, which in turn determines its practical applications. We make a case for an alternative approach to grouping synthetic data types that better reflect privacy perspectives in order to facilitate regulatory guidance in the generation and processing of synthetic data. This approach to classification provides flexibility to new advancements like deep generative methods and offers a more practical framework for future applications.

## Introduction

AI development is data hungry and creating synthetic data - data that is "artificially generated rather than captured in the real world" - is becoming increasingly attractive as a privacy conscious and potentially cost-effective way forward, endorsed by the European Data Protection Supervisor (EDPS).[1-3] Generation method choices will influence several ethical concerns like privacy, bias and fairness and computational complexity. Privacy is of particular interest to if, how or when the synthetic data can be used according to data protection regulations and highly dependent on generation method.

Synthetic data can either be created based on existing knowledge (e.g. manually or from simulations) or it can be created from data captured from real-world events ("real data"). Synthetic data created from existing knowledge rather than real data, would have a negligible risk of disclosure of sensitive data. For synthetic data that is derived from real data, the method of generation will influence how robust the dataset is for reverse engineering.[4]

The level of privacy risk in a dataset has implications for how it can be processed legally, and how it should be protected. This has very specific practical consequences for its use and sharing.[5] As the techniques for identification have become more sophisticated and new methods for synthetic data generation have evolved, there is a need for a more granulated grouping of synthetic data types that reflect the residual privacy risk (see Figure 1).

The division of synthetic data types that we endorse is based broadly on the generation method and how or if these are derived from real data, making it easy to implement and understand.

[1] *University of Oslo*
[2] *DNV*
[3] *UiT, The Arctic University of Norway*
[4] *Cancer Registry of Norway, Norwegian Institute of Public Health*

This more intuitive taxonomy incorporates consideration of privacy risk level, which will aid in establishing guidelines for understanding regulatory constraints, thus providing a practical framework and communication tool for increasing the understanding between synthetic data users and regulators. Potential fairness and bias amplifications are crucial considerations in healthcare datasets generated synthetically and are reliant on the original training data and preprocessing methods.

## Traditional grouping and its insufficiencies

A common way of grouping synthetic data in the literature is according to a three-way split: *fully synthetic datasets*, *partially synthetic datasets* and *hybrid data*. [6] This definition can be traced to the 2017 paper "A Review Of Synthetic Data Generation Methods For Privacy Preserving Data Publishing" by Surendra and Mohan [7]. The authors define *partially synthetic* data as datasets where variables with a high risk of disclosure are replaced by synthetic values. They introduce *fully synthetic data* as a synthetic dataset where all variables are synthetic and "randomly drawn from estimated density functions", examples being multiple imputation and bootstrap methods. *Hybrid data* is explained as combining real and synthetic records to form hybrid data.

The three "types" are thus different in nature – partially and fully synthetic data are defined based on the composition of synthetic vs real samples in the dataset, and hybrid data is based on a specific method for generating synthetic samples where a real and a synthetic sample is combined to create a hybrid dataset

One could imagine a partially synthetic dataset where the synthetic samples were made by hybrid masking. Consequently, the "types" are not mutually exclusive but overlapping. The "types" can also be said to not be collectively exhaustive, as manually created synthetic data or other knowledge-based methods are not covered by either of the definitions.

For this three-way split, the paper references a book from 2008: "*Privacy Preserving Data Mining: Models and Algorithms",* edited by Aggarwal and Yu.[8] In this book, however, there was no attempt to create a taxonomy nor was a three-way split of synthetic data types introduced. The book reviews examples of techniques for creating synthetic data that were relevant at the time: Multiple imputation, Bootstrap, Latin hypercube sampling, Cholesky Decomposition and hybrid masking. In the example of Cholensky Decomposition, sensitive variables were synthesized to create a partially synthetic dataset, but there was no mention of the concept of fully synthetic datasets. The Surendra & Mohan types of synthetic data are still widely referenced, as for example in Gonzales, et al. [6] 2023 review paper.

Murtaza et al used a different approach in their review paper in 2023, where the mix of the dataset – its "syntheticity" was defined as either partially synthetic or fully synthetic, but hybrid data was not mixed into the definitions.[9] The authors continue to propose a division between *approaches* of synthetic data generation: Knowledge driven, Data driven (Transformational and Simulations) and Hybrid. The term *Simulation* is used to describe machine learning based generative models, rather than simulation models that are built manually, for example based on physical constraints. In their paper, the authors do not use the term "types" of synthetic data specifically nor do they propose this as a new classification. Although being superior in its logic, an investigation of references suggests their definition has not caught on widely.[10]

The division between a partially synthetic dataset – a mixed dataset - and a fully synthetic dataset is a practical distinction if not limited to a specific generation method. The original idea of the division from 2017 was based on that datasets containing real data have higher disclosure risk, and the fully synthetic had negligible risk yet also low utility.[7]

In recent years, the literature has seen the rise of ML and deep generative methods that have made it possible to create fully synthetic datasets with high utility and a relatively low disclosure risk. Additionally, the assumption that fully synthetic data has a strong privacy protection since the released data is completely artificially generated and doesn't contain original data has since been discredited.[11,12] Risk of identity disclosure will depend on the generation method and characteristics of the data itself, also when all variables are synthetic.[13] This renders a need for a more granulated approach in grouping of synthetic data.

## Precedent in the AI ecosystem

Despite the widespread uptake of the Surendra and Mohan definitions, it is more common in the field of machine learning (ML) to taxonomize according to methodology as opposed to outcome. For example, if we consider generative AI and supervised and unsupervised learning – the process is described, as opposed to the outcome. If the outcome was utilized for taxonomy purposes, an overlap would arise with other learning methodologies that produce optimized behaviors e.g. reinforcement learning, blurring the distinction between the groups. This situation reflects the present state observed in traditional synthetic data classification.

## Methodological classification of synthetic data

This suggested taxonomy builds on and amends the thinking introduced by Murtaza et al. , where the methodology dictates the data type.[9] The first division is between knowledge-based and data driven synthetic data as illustrated in Figure 1.

### Knowledge-based synthetic data

**Knowledge-based synthetic data** can be either created manually or by simulation of a model. New synthetic data can be manually created, for example when a clinician creates examples of typical entries in a patient record based on years of experience. Simulation-based or rule-based data generation replicates real-world processes to produce synthetic datasets, for example when modeling a digital twin of coronary heart vessels.[14] Simulating events or behaviors based on established rules and variables enables the examination of interactions within dynamic systems. This approach is frequently employed to test scenarios and evaluate potential outcomes, offering valuable insights into complex systems such as customer service workflows, supply chains, and healthcare operations. These knowledge-based examples have no relationship with any real datapoint and therefore in principle should have a negligible disclosure risk. In terms of disclosure risk, one could imagine a situation where an expert remembers a real case and inadvertently copies this when manually creating synthetic data. This would be a case of a breach of confidentiality. Similarly, the risk of bias propagation would depend on the experiences used to build the data. Computational carbon footprint may be low, but the generation could be highly dependent on manual labor.

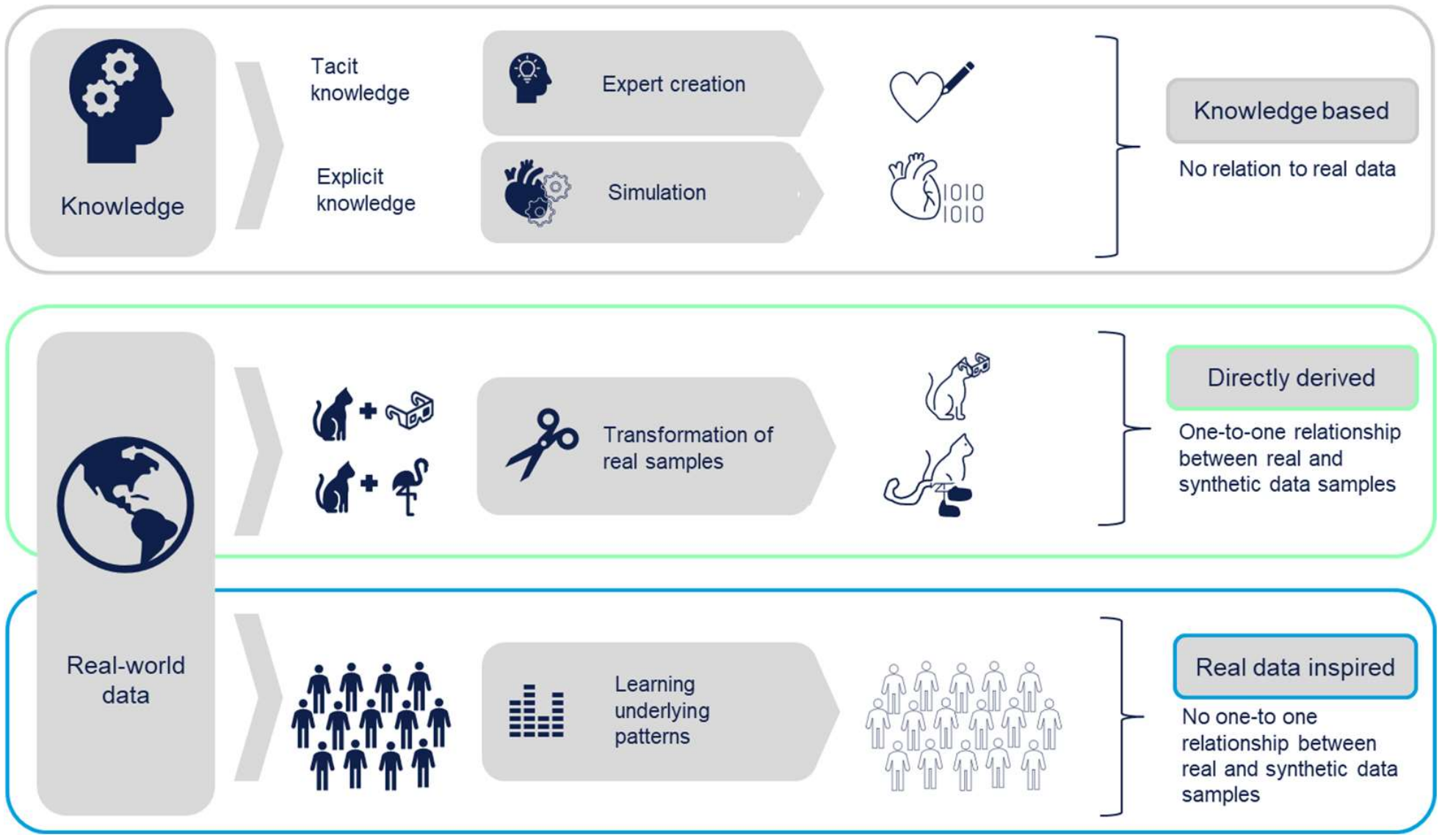

*Figure 1 A division of synthetic data into distinct groups according to methodology of generation.*

## Data-driven synthetic data

Synthetic data generation that is based on real data has evolved significantly over the past decades, progressing through distinct technological generations.[15] Early approaches relied on modifying real samples by for example adding errors, using rule-based algorithms for modifying existing datasets, before advancing to more sophisticated interpolation-based methods like SMOTE and ADASYN that could generate new samples through neighborhood relationships. [15] Increasingly complex statistical methods were developed. The field underwent a revolutionary shift with the introduction of ML models and subsequent deep generative approaches such as GANs and VAEs [16], which learn the underlying data distribution to create realistic synthetic samples. These newer approaches, while being more complex to implement [17], offer powerful capabilities for generating high-quality synthetic data that closely mirrors real-world distributions.[3,18] While certain methods can mitigate the potential augmenting of biases during generation, bias risk in data driven synthetic data would depend highly on the characteristics of the original training dataset. The carbon footprint tends to be significantly higher for the more advanced deep learning-based generation methods.

To summarize, the techniques are either "inspired" by the real data, learning the underlying structures of the data or they are directly derived from the real datapoints. When real data samples are transformed - either distorted or used as a basis for a calculation to create a new synthetic point, we can call them directly or "one-to-one" derived from real data. This is in line with the Murtaza concept of Transformational synthetic data.[9] The direct relationship between original and synthetic datapoints may make the datasets more vulnerable to reverse engineering and inferencing. Examples of techniques to create ***Directly derived synthetic data*** are hybrid masking, noise perturbations, scaling and rotation transformations.

When an ML model learns the underlying statistical patterns of a dataset and uses this to generate new data with no one-to-one relationship with the real datapoints, we can call these ***Real data inspired synthetic data.*** This definition is close to the type called *Simulations* by Murtaza et al.[9] While ML based methods most certainly are of this group, techniques like SMOTE [19] are more challenging to classify. It could be called real-world data *inspired* as it is based on interpolation and therefore a many-to-one relation, while in the Murtaza view it may not fit within the definition of type *Simulations*.

The disclosure risk varies not only depending on the method used but also characteristics of the data. Even when there is no one-to-one relationship between the data samples, generative models may overfit and create samples that are very similar or exact matches to real samples leaving it vulnerable to inferencing.[20] The risk – although likely lower in real data inspired synthetic data, should not automatically be assumed zero. Tailored approaches to assess the risks for each of the three groups may be appropriate, and so this logical classification can provide a basis for supporting disclosure risk assessment and the accompanying regulatory documentation if required.

As suggested by Murtaza, hybrid or mixed versions of approaches to synthetic data types could create alternative mixed types of synthetic data but do not necessarily require a separate classification, as they inherit properties of their original families.

Datasets could be either partially synthetic or fully synthetic, regardless of the method used to generate the synthetic data. Inspired by the field of federated learning, one could imagine a partially synthetic dataset where real data points are replaced either vertically (feature-wise) or horizontally (record-wise).[21] The disclosure risk in horizontal datasets would naturally be higher, as certain records are left unchanged.

Table 1 summarizes a comparative validation of the taxonomies discussed.

*Table 1 Comparative validation of the different suggested taxonomies of synthetic data types.*

| | **Composition** | **Complete?** | **Mutually exclusive?** | **Hierarchical clarity?** | **Practical?** |
|---|---|---|---|---|---|
| Surendra & Mohan[7] | • Fully synthetic<br>• Partially synthetic<br>• Hybrid data | ✗<br>No, does not cover SD derived from knowledge | ✗<br>No, hybrid data could be used for a partially synthetic dataset. | ✗<br>No, partial and fully synthetic data are not same level as hybrid data (sample level versus dataset level). | ÷<br>Newer techniques go into the fully synthetic data group. This leaves very little granularity. |
| Murtaza et al.[9] | • Knowledge driven<br>• Data driven (Transformational and Simulations)<br>• Hybrid (part knowledge, part data driven) | ✓ | ÷<br>Hybrid is a mixed group and could overlap. | ✓ | ÷<br>Aligned with newer methods, may need further granularity in the ML group.<br>The term “Simulation” of the ML group may create confusion. |
| Amended Murtaza et al. (this suggestion) | • Knowledge based<br>• Directly derived<br>• Real data inspired | ✓ | ✓ | ✓ | ÷<br>Aligned with newer methods, may need further granularity in the ML group. |

## Conclusion

We suggest conducting classifications of synthetic data types based on generation methods building on the suggestion of Murtaza et al. in order to replace, and also increase granularity of the traditional output derived division of “fully, partial and hybrid” synthetic data:

**Knowledge based**: synthetic data that is based on either tacit or explicit knowledge and created either manually by an expert or simulated.

**Data driven, divided into:**

**Directly derived**: synthetic data that is based on real-world data and has a one-to-one relationship between the real and synthetic samples.

**Real data inspired**: synthetic data that is based on real-world data but does not have a one-to-one relationship between the real and synthetic samples.

This division acknowledges the precedent within the AI ecosystem where the processing method determines taxonomy. It also supports downstream consideration of the residual disclosure risk in a dataset resulting from the underlying generation methods.

The concept of a fully synthetic dataset versus a partially synthetic dataset is still valid and could be useful but is not complete to describe the different flavors of synthetic data we see today. The concepts are orthogonal to the groups outlined in this paper, as both fully and partially synthetic datasets can be constructed using any of the three types of synthetic data: knowledge based, directly derived, or real data inspired.

This updated approach to categorizing synthetic data types integrates privacy considerations. Agreeing on more practical classifications represents a critical step to providing clearer regulatory guidance for the processing of synthetic data. Furthermore, this classification framework accommodates emerging technologies such as deep generative methods, offering greater flexibility and a more practical foundation for future applications.